%% file: top.tex
\documentclass{article}

\usepackage[preprint]{corl_2022} 

\usepackage{graphicx}
\usepackage{comment}
\usepackage{amsmath,amsfonts,amssymb} 
\usepackage[utf8]{inputenc}
\usepackage{epsfig}
\usepackage{graphicx}
\usepackage[nice]{nicefrac}
\usepackage{caption}
\usepackage{subcaption}
\usepackage{xspace}
\usepackage{array}
\usepackage[nolist, nohyperlinks]{acronym}
\usepackage{booktabs}
\usepackage{setspace}

\input{shortcuts}
\input{acronyms}

\title{\methodName{}: Generating Safety-Critical Driving Scenarios\\ for Robust Imitation via Kinematics Gradients}

\author{Niklas Hanselmann\textsuperscript{1,2}\hspace{24pt}%
Katrin Renz\textsuperscript{2,3}\hspace{24pt}%
Kashyap Chitta\textsuperscript{2,3}\hspace{24pt}\\[0.5ex]%
\textbf{Apratim Bhattacharyya\textsuperscript{2,3}\hspace{24pt}}%
\textbf{Andreas Geiger\textsuperscript{2,3}\hspace{8pt}}\\[2ex]
\textsuperscript{1}Mercedes-Benz AG R\&D, Stuttgart\hspace{24pt}%
\textsuperscript{2}University of Tübingen\\
\textsuperscript{3}Max Planck Institute for Intelligent Systems, Tübingen}

\begin{document}
\maketitle
\setstretch{1.08} 

\input{sec_abstract}


\input{sec_intro}
\input{sec_related}
\input{sec_method}
\input{sec_results}
\input{sec_conclusion}

\acknowledgments{This work was supported by the German Federal Ministry for Economic Affairs and Climate Action within the project KI Delta Learning (project numbers: 19A19013A, 19A19013O), the German Federal Ministry of Education and Research (Tübingen AI Center, FKZ: 01IS18039A, 01IS18039B) and the German Research Foundation (SFB 1233, Robust Vision: Inference Principles and Neural Mechanisms, TP 17, project number: 276693517). We thank the International Max Planck Research School for Intelligent Systems (IMPRS-IS) for supporting Katrin Renz and Kashyap Chitta. The authors also thank Aditya Prakash and Bernhard Jaeger for proofreading.}
\clearpage
\bibliography{bibliography_long,bibliography,bibliography_custom}  

\end{document}

%% file: shortcuts.tex
\newcommand{\ba}{\mathbf{a}}

\newcommand{\bo}{\mathbf{o}}

\newcommand{\bs}{\mathbf{s}}

\newcommand{\bw}{\mathbf{w}}
\newcommand{\bx}{\mathbf{x}}

\newcommand{\btheta}{\boldsymbol{\theta}}

\newcommand{\nE}{\mathbb{E}}

\newcommand{\nR}{\mathbb{R}}

\newcommand{\cD}{\mathcal{D}}

\newcommand{\cL}{\mathcal{L}}
\newcommand{\cM}{\mathcal{M}}

\newcommand{\cR}{\mathcal{R}}
\newcommand{\cS}{\mathcal{S}}

\newcommand{\figref}[1]{Fig.~\ref{#1}}
\newcommand{\secref}[1]{Section~\ref{#1}}

\newcommand{\tabref}[1]{Table~\ref{#1}}

\DeclareMathOperator*{\argmin}{argmin~}

\makeatletter
\DeclareRobustCommand\onedot{\futurelet\@let@token\@onedot}
\def\@onedot{\ifx\@let@token.\else.\null\fi\xspace}
\def\eg{e.g\onedot} 
\def\ie{i.e\onedot}

\def\wrt{wrt\onedot}

\makeatother

\newcommand{\boldparagraph}[1]{\vspace{0.2cm}\noindent{\bf #1:} }

\definecolor{darkgreen}{rgb}{0,0.7,0}

\newcommand{\carla}{CARLA}
\newcommand{\methodName}{KING} 
\newcommand{\bevAgent}{{AIM-BEV}}

\newcommand{\nhnotetoself}[1]{}

\definecolor{ellisred}{rgb}{0.87,0.44,0.38} 
\definecolor{ellisgreen}{rgb}{0.69,0.90,0.52} 
\definecolor{elliscyan}{rgb}{0.29,0.77,0.74} 
\definecolor{ellisorange}{rgb}{0.89,0.55,0.28} 
\definecolor{ellisblue}{rgb}{0.41,0.61,0.86} 

\newcommand{\ellisred}[1]{\noindent{\color{ellisred}{#1}}}

\newcommand{\ellisblue}[1]{\noindent{\color{ellisblue}{#1}}}

%% file: acronyms.tex
\begin{acronym}
    \acro{bbo}[BBO]{black-box optimization}
\end{acronym}

\begin{acronym}
    \acro{il}[IL]{imitation learning}
\end{acronym}

\begin{acronym}
    \acro{rl}[RL]{reinforcement learning}
\end{acronym}

\begin{acronym}
    \acro{bev}[BEV]{bird's-eye view}
\end{acronym}

\begin{acronym}
    \acro{gail}[GAIL]{generative adversarial imitation learning}
\end{acronym}

\begin{acronym}
    \acro{gru}[GRU]{gated recurrent unit}
\end{acronym}

\begin{acronym}
    \acro{rc}[RC]{Route Completion}
\end{acronym}

\begin{acronym}
    \acro{is}[IS]{Infraction Score}
\end{acronym}

\begin{acronym}
    \acro{bo}[BO]{Bayesian Optimization}
\end{acronym}

\begin{acronym}
    \acro{cvae}[CVAE]{conditional variational autoencoder}
\end{acronym}

%% file: sec_abstract.tex
\begin{abstract}
    Simulators offer the possibility of safe, low-cost development of self-driving systems. However, current driving simulators exhibit naïve behavior models for background traffic. Hand-tuned scenarios are typically added during simulation to induce safety-critical situations. An alternative approach is to adversarially perturb the background traffic trajectories. In this paper, we study this approach to safety-critical driving scenario generation using the CARLA simulator. We use a kinematic bicycle model as a proxy to the simulator's true dynamics and observe that gradients through this proxy model are sufficient for optimizing the background traffic trajectories. Based on this finding, we propose KING, which generates safety-critical driving scenarios with a 20\% higher success rate than black-box optimization. By solving the scenarios generated by KING using a privileged rule-based expert algorithm, we obtain training data for an imitation learning policy. After fine-tuning on this new data, we show that the policy becomes better at avoiding collisions. Importantly, our generated data leads to reduced collisions on both held-out scenarios generated via KING as well as traditional hand-crafted scenarios, demonstrating improved robustness.
\end{abstract}

%% file: sec_intro.tex
\section{Introduction}
After years of steady progress, autonomous driving systems are getting closer to maturity~\cite{Janai2020}. Due to the high consequences of failure, they have to satisfy extraordinarily high standards of robustness in the face of unseen and safety-critical scenarios. However, real-world data collection and validation for these situations is dangerous and lacks the necessary scalability~\cite{Okelly2018NEURIPS,Norden2019ARXIV}. These problems can be addressed with realistic simulation. Unfortunately, current simulators such as CARLA \cite{Dosovitskiy2017CORL} are not only insufficient in terms of visual fidelity but also lack the necessary diversity of driving scenarios: there exists both an \emph{appearance} and a \emph{content gap} to the real world~\cite{Kar2019ICCV}. The content gap poses a major challenge in the adoption of driving agents trained in simulation using \ac{il} or \ac{rl}, which are often brittle to o.o.d. inputs underrepresented during training~\cite{Filos2020ICML}. In this work, we aim to address the content gap by improving the behavior of simulated background traffic agents.

\begin{figure}
    \centering
    \includegraphics[width=\linewidth]{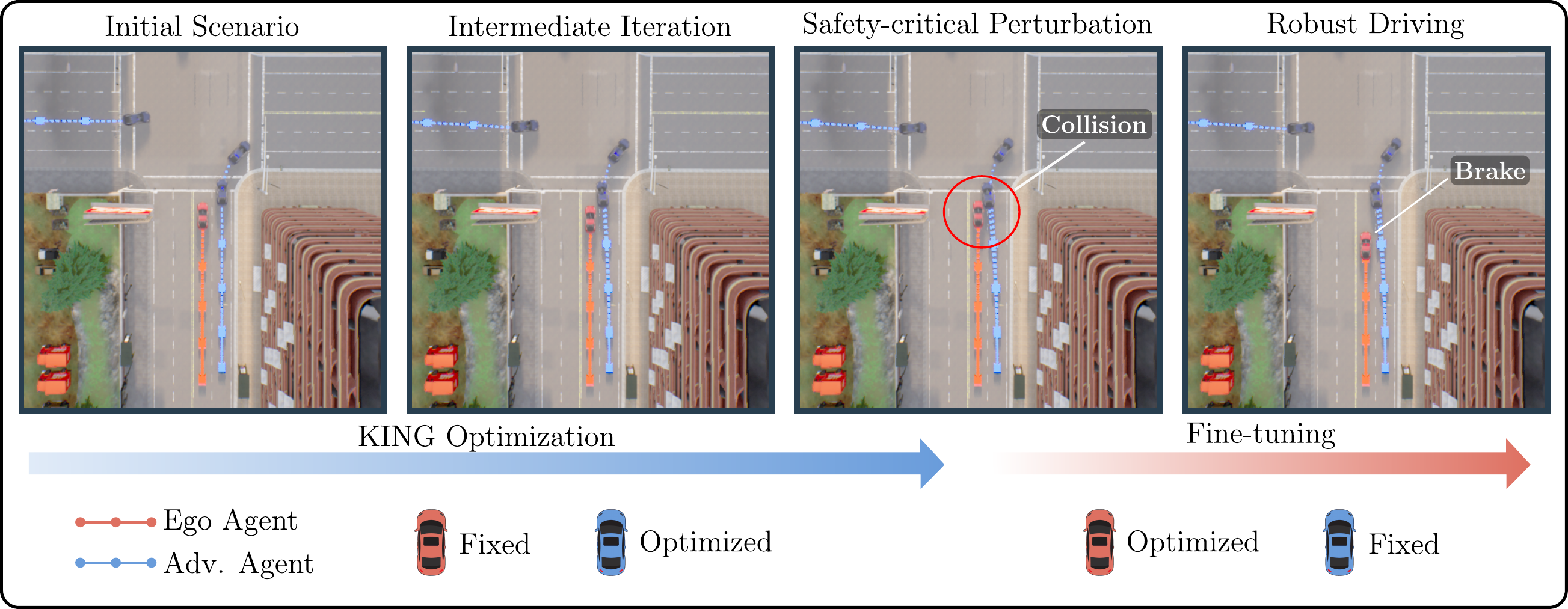} \\
    \caption{\textbf{Generating safety-critical scenarios for robust driving.} Left: we propose \methodName{}, a novel optimization method to generate safety-critical driving scenarios which iteratively updates the initial scenario using gradients through a differentiable kinematics model and successfully induces a collision with the ego agent. Right: fine-tuning on expert behavior in safety-critical perturbations leads to a more robust agent.
    }
    \label{fig:teaser}
\end{figure}

Background agents in current simulators follow naïve behavioral models, resulting in limited diversity of the emerging traffic~\cite{Dosovitskiy2017CORL,Richter2017ICCV}. Critical scenarios are often hand-crafted~\cite{CarlaChallenge2021,Fremont2020ARXIV}. This strategy is unlikely to be successful in fully covering the long-tailed distribution of critical situations that might be encountered in the real world. Furthermore, these scenarios are often non-adaptive to the driving agent under test. A more targeted approach is to actively seek possible failure modes. To do so, existing work perturbs the trajectories of background agents in a physically plausible manner to induce failures in the driving agent \cite{Abeysirigoonawardena2019ICRA, Wang2021CVPRb}. This paradigm can be framed as a kinematically constrained adversarial attack on the driving agent, where the amount of safety-critical data generated within a given compute budget is dependent on the success rate of the attack. The prevalent approach for this task is \ac{bbo}, since simulators are often not differentiable~\cite{Guo2019ICML, Ilyas2019ICLR}.  However, as we observe on the widely used \carla{} simulator~\cite{Dosovitskiy2017CORL}, existing attacks based on \ac{bbo} (\eg ~\cite{Guo2019ICML,Hansen2001ecj,Ilyas2019ICLR}) struggle to reliably induce collisions in \ac{il}-based driving agents (see \tabref{tab:num_agents}).

As observed in image-space adversarial attacks, gradient-based optimization has the potential to be faster and more successful than \ac{bbo}~\cite{Andriushchenko2020ECCV,Deng2020ARXIV}.
Moreover, there has been a trend towards end-to-end differentiability, both in simulation \cite{Scibior2021ARXIV, Scheel2021CORL, Bergamini2021ICRA, Suo2021CVPR} and driving agents \cite{Bojarski2016ARXIV, Sadat2020ECCV, Casas2021CVPR, Codevilla2018ICRA, Prakash2020CVPR, Chitta2021ICCV}.
Using differentiable components enables {gradient-based generation of adversarial traffic scenarios}.
In this paper, we answer an important question: \textit{does the entire simulation pipeline need to be differentiable to provide useful gradients for the optimization of traffic scenarios?} We present \methodName{}, a simple and effective approach for safety-critical scenario generation. Our key idea is to use a kinematic bicycle model as proxy to a driving simulator's true dynamics, and solve for safety-critical perturbations of non-critical initial scenarios via backpropagation. The process of optimizing a non-critical scenario with \methodName{} is visualized in \figref{fig:teaser} (left). Further, we show that \methodName{} generates challenging but solvable test cases for driving systems that use both (1) a planner that acts on a \ac{bev} grid input and (2) a camera and LiDAR-based driving agent~\cite{Prakash2021CVPR}. Finally, we demonstrate that scenarios generated by \methodName{} can augment the original training distribution which has limited diversity. This leads to improved collision avoidance, as shown in \figref{fig:teaser} (right).

\boldparagraph{Contributions} 
(1) We propose \methodName{}, a simple procedure for generating safety-critical scenarios via backpropagation that is more reliable and requires less optimization time than \ac{bbo}. (2) We show that \methodName{} generates challenging, diverse, and solvable scenarios for two different driving agents with different input modalities. (3) We use the generated scenarios to augment the CARLA simulator's non-diverse traffic, improving the robustness of an end-to-end \ac{il}-based driving agent on both our generated test cases and a benchmark containing CARLA's hand-crafted scenarios. Project page: \url{https://lasnik.github.io/king/}.

%% file: sec_related.tex
\section{Related Work}
\label{sec:related}

\boldparagraph{End-to-End Driving} We are interested in stress-testing and improving end-to-end learning-based autonomous driving systems. While there are a few \ac{rl} methods for this task \cite{Chen2021ICCVb, Toromanoff2020CVPR}, most work leverages \ac{il}. Some adhere closely to the end-to-end learning paradigm \cite{Pomerleau1988NIPS,Bojarski2016ARXIV, Codevilla2018ICRA, Codevilla2019ICCV, Ohn-Bar2020CVPR, Prakash2021CVPR}, directly inferring driving actions from raw sensor observations. However, others use interpretable intermediate representations~\cite{Sauer2018CORL,Xiao2020CORL,Behl2020IROS}. In particular, \ac{bev} semantic occupancy grid representations are widely used in modern driving approaches~\cite{Zhou2019CORL, Sadat2020ECCV, Casas2021CVPR, Zhang2021ICCV, Chitta2021ICCV}. 
This representation can be inferred from images~\cite{Mani2020WACV,Roddick2020CVPR,Pan2020RAL,Hendy2020ARXIV,Hu2021ICCV,Chitta2021ICCV,Philion2020ECCV,Loukkal2020ARXIV}. In our study we consider two \ac{il}-based driving agents reflecting both schools of thought: (1) a planner called AIM-BEV acting on ground-truth perception represented as a \ac{bev} semantic occupancy grid, and  (2) an end-to-end agent acting on camera and LiDAR observations called TransFuser~\cite{Prakash2021CVPR}.

\boldparagraph{Generating Safety-Critical Scenarios} Previous work on generating safety-critical scenarios relies on \ac{bbo} techniques and explores a variety of search space parameterizations, such as initial velocity and position of a single adversarial agent \cite{Ding2020IROS, Ding2021RAL}, a high-level route graph \cite{Abeysirigoonawardena2019ICRA} or sampling weights for the final layer of a driving policy from an ensemble \cite{Okelly2018NEURIPS}. In AdvSim \cite{Wang2021CVPRb}, the search space is parameterized as a sequence of kinematic bicycle model states for each adversarial agent, with steering and acceleration actions as free parameters. We also adopt this simple and expressive parameterization for \methodName{}. Different from this line of work, we propose a gradient-based procedure to optimize over these parameters rather than resorting to \ac{bbo} techniques.
Concurrent work presents STRIVE \cite{Rempe2021ARXIV}, a framework that also generates critical scenarios via gradient-based optimization. Here, an adversarial agent is parameterized as a latent vector of a learned motion forecasting model. While they only attack a simple, privileged rule-based planner, we focus on end-to-end \ac{il} agents. Furthermore, we empirically compare \methodName{} to black-box scenario generation, which is not considered in STRIVE. Lastly, STRIVE uses a proxy of the driving agent to enable gradient-based optimization, while \methodName{} directly optimizes for collisions wrt. the actual driving agent.

%% file: sec_method.tex
\section{Safety-Critical Scenario Generation for Robust Imitation}
\label{sec:method}

\begin{figure}[t!]
\centering
\includegraphics[width=\columnwidth]{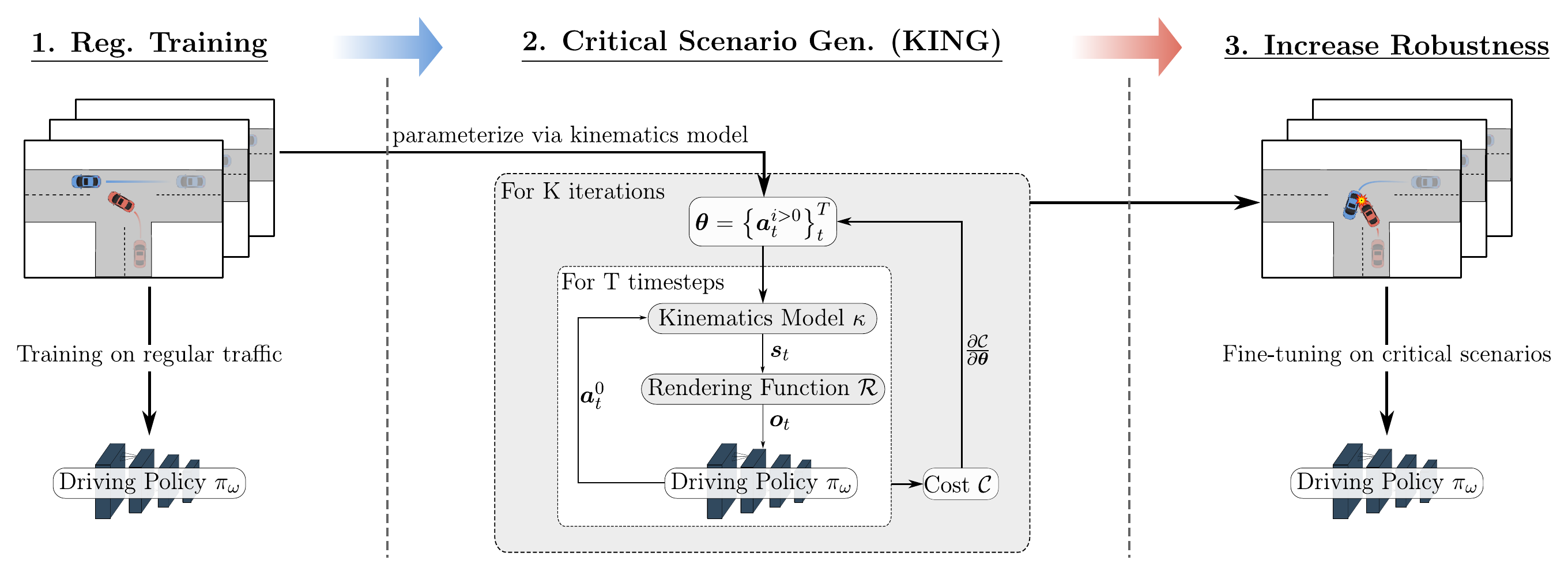}
\caption{\textbf{Robust training pipeline.} Given any agent with a driving policy $\pi_{\omega}$ trained on regular traffic data, we propose to increase its robustness under safety-critical scenarios by generating targeted augmentations. We propose \methodName{}, a gradient-based optimization procedure to obtain safety-critical perturbations of initial regular traffic scenarios. These perturbations then serve as additional training data for $\pi_{\omega}$.}
\label{fig:pipeline}
\vspace{-1em}
\end{figure}

In this section, we outline our overall approach for stress-testing and improving the robustness of \ac{il}-based driving agents, which is illustrated in \figref{fig:pipeline}. Given a driving agent trained on regular traffic, we propose \methodName, a novel gradient-based optimization procedure for automatically generating safety-critical perturbations of non-critical scenarios tailored to the agent under consideration. These scenarios serve to augment the original training distribution with limited diversity. In the following, we formally present our task settings, detail the parameterization and objective function used for scenario generation, and describe our robust training approach for \ac{il}.

\boldparagraph{Driving Agent and Regular Training}
\label{sec:method:problem}
We are interested in stress-testing and improving the robustness of an \ac{il}-based driving agent trained on a dataset $\mathcal{D}_{reg}$ of expert driving in regular traffic. We assume that the driving policy of the agent is a neural network $\pi_\omega$ with parameters $\omega$ that takes in an observation $\bo_t \in \nR^{H_o \times W_o \times C_o}$ and goal location $\bx_{goal} \in \nR^2$ indicating the intended high-level route on the map, and plans a trajectory represented by four future 2D waypoints $\bw \in \nR^{4 \times 2}$:
\begin{equation}
    \pi_\omega \left(\bo_t, \bx_{goal} \right) : \nR^{H_o \times W_o \times C_o} \times \nR^2 \to \nR^{4 \times 2}.
    \label{eq:method:policy}
\end{equation}

Based on the predicted waypoints, the final actions $\ba_t^{0} \in \left[-1, 1\right]^2$ in the form of throttle and steering commands are produced by lateral and longitudinal controllers. Currently, several state-of-the-art \ac{il} agents fall under this paradigm~\cite{Chen2019CORL,Prakash2021CVPR,Chitta2021ICCV}. With this general scheme, we consider both an \ac{il} policy with an intermediate representation as well as a strictly end-to-end model in our study. The first is a planner acting on ground-truth visual abstractions which we will refer to as \bevAgent{}. This is inspired by \cite{Behl2020IROS} and the AIM-VA model in \cite{Chitta2021ICCV}, but uses a \ac{bev} intermediate representation instead of 2D semantics since the \ac{bev} is an orthographic projection of the physical 3D space which is better correlated with vehicle kinematics than the 2D image domain. Here, the observations $\bo_t \in \nR^{192 \times 192 \times 3}$ are a rasterized \ac{bev} grid encoding HD map information with channels for (1) road and (2) lanes as well as a separate channel for dynamic obstacles such as background agents (3). The grid represents the environment ahead and to each side of the agent at a resolution of 5 pixels per meter. In addition to \bevAgent{}, we also stress-test the publicly available checkpoint\footnote{\url{https://github.com/autonomousvision/transfuser/tree/main/transfuser}} released by the authors of TransFuser~\cite{Prakash2021CVPR}. This is a recent state-of-the-art IL-based self-driving model acting on observations $\bo_t^{rgb} \in \nR^{256 \times 256 \times 3}$ obtained from a front-facing camera and a discretized \ac{bev} lidar-histogram with two height bins $\bo_t^{lid} \in \nR^{256 \times 256 \times 2}$. Both \bevAgent{} and TransFuser are trained on observation-waypoint pairs $\left(\bo, \bw\right)$ drawn from $\mathcal{D}_{reg}$. The observations are mapped to a latent representation which is input to a \ac{gru} that plans the trajectory $\bw$ in an autoregressive fashion. For additional details, please refer to the supplementary material and the original TransFuser paper.

\boldparagraph{Gradient-based Scenario Generation}
To optimize for safety-critical perturbations of an initial non-critical scenario (regular traffic), we iteratively simulate the scenario with the driving agent under attack (ego agent) in a closed-loop simulation. In particular, we aim to create a collision between the ego agent and one of the background actors (adversarial agents). At each iteration, we adjust the scenario's parameters (\ie the trajectories of adversarial agents) in order to induce such a collision. Importantly, the ego agent is able to react to the perturbations of the adversarial agents, since the attacks take place in a closed loop. Therefore, the scenarios generated are adaptive to the specific ego agent being attacked. In the following, we formally describe the simulation process and scenario generation procedure.

Let $\bx_t^i \in \nR^2$, $\psi_t^i \in \left[0, 2\pi\right]$ and $v_t^i \in \nR$ be the ground-plane position, orientation and speed of the $i$-th agent at time $t$, where the index 0 indicates the ego agent. We denote the traffic state as $\bs_t = \left\{\bx_t^i, \psi_t^i, v_t^i \right\}_{i=0}^N$, where $N$ is the number of agents. In slight abuse of notation, we will use $\bs_t^i$ to refer to the state of a specific agent. We instantiate a particular scenario as a sequence of these states $\cS = \left\{\bs_t\right\}_{t=0}^T$, where $T$ is a fixed simulation horizon. $\cS$ is initialized using regular, non-critical traffic behavior as described in \secref{sec:results:bbo}. To unroll the simulation forward in time, we compute the state at the next timestep $\bs_{t+1}$ given the current state $\bs_{t}$ and actions of all agents $\ba_t = \left\{\ba_{t}^i\right\}_{i=0}^N$ using the kinematics model $\kappa$, \ie, $\bs_{t+1} = \kappa \left(\bs_t, \ba_t \right)$. We choose the bicycle model, which provides a strong prior on physically plausible motion of non-holonomic vehicles \cite{Polack2017IV, Chen2021ICCVb} and is differentiable, enabling backpropagation through the unrolled state sequence $\cS$. The ego agent is reactive to the simulation and chooses its actions $\ba_t^0$ based on observations $\bo_t$ of the true underlying state, which are obtained through a rendering function $\mathcal{R}$, \ie, $\bo_t = \cR \left(\bs_t, \cM\right)$. To render \ac{bev} semantic occupancy grids for \bevAgent, we query a differentiable rasterizer \cite{Jaderberg2015NIPS} for the given current state $\bs_t$ and HD map $\mathcal{M}$, representing other agents by their bounding polygons. To render sensor data such as camera imagery and LiDAR point clouds for TransFuser, we query the CARLA simulator's graphics engine. Note that all components of the simulation ($\pi$, $\kappa$ and $\mathcal{R}$) are differentiable for \bevAgent{} but $\mathcal{R}$ is not differentiable for TransFuser.

\boldparagraph{Safety-Critical Perturbation}
We perturb the sequence of states $\left\{\bs_t^{i>0}\right\}_{t=0}^T$ for the $N$ adversarial agents in order to induce a collision. If the ego agent collides within $T$ timesteps, we terminate the simulation successfully. At the same time, we would like the behavior of the adversarial agents to remain plausible. If any adversarial agent deviates from the drivable areas of the map or collides with another adversarial agent, the simulation terminates unsuccessfully. To detect collisions, we perform intersection checks between the bounding boxes of the agents. For out-of-bounds violations, we check if the adversarial agent bounding boxes enter the off-road area of the map.

Similar to \cite{Wang2021CVPRb}, we parameterize the trajectories of adversarial agents as a sequence of states obtained by unrolling the kinematics model $\kappa$. Specifically, a safety-critical perturbation is found by optimizing the sequence of actions $\left\{\ba_t^{i>0}\right\}_{t=1}^T$ for each adversarial agent. The overall search space can be written as $\btheta = \left\{\btheta^i \right\}_{i=1}^N$ where $\btheta^i = \left\{\ba_{t=0}^i, ..., \ba_{t=T}^i \right\}$, with dimensionality $N\times T\times2$. 
We optimize an objective $\mathcal{C}$ which is motivated by prior work on safety-critical scenario generation~\cite{Abeysirigoonawardena2019ICRA,Ding2020IROS,Wang2021CVPRb}:
\begin{equation}
    \btheta^\ast = \argmin_{\btheta} \mathcal{C}(\cS) \quad \text{with } \quad \mathcal{C}(\cS) =\phi_{col}^{ego}(\cS) + \lambda \ \phi_{col}^{adv}(\cS) + \gamma \ \phi_{dev}^{adv}(\cS).
    \label{eq:method:objective}
\end{equation}    
We encourage collisions involving the ego agent with the cost $\phi_{col}^{ego}$ and discourage unsuccessful terminations of the simulation via the costs $\phi_{col}^{adv}$ and $\phi_{dev}^{adv}$, weighted using hyper-parameters $\lambda$ and $\gamma$. These costs are similar to commonly used cost functions in planning~\cite{Zeng2019CVPR, Sadat2020ECCV, Casas2021CVPR}. We now explain the costs of the objective $\mathcal{C}$ in detail.

Let $d(\bs^i_t,\bs^j_t)$ denote the Euclidean distance in $\mathbb{R}^2$ between closest points on the bounding polygons of the $i^{\text{th}}$ and $j^{\text{th}}$ agents at time $t$. To induce failure in the driving policy $\pi_{\omega}$ through a collision with an adversarial agent, we choose $\phi_{col}^{ego}$ to be an attractive potential encouraging close encounters. We minimize the Euclidean distance between the ego agent and closest adversarial agent. We choose only the closest adversarial agent in order to discourage situations where multiple adversaries deviate from their trajectory to collide with the ego agent:
\begin{equation}
    \phi_{col}^{ego}(\cS) = 
         \min_{i \in \left\{1, ..., N\right\}} \frac{1}{T} \sum\limits_{t=0}^{T} d(\bs^0_t,\bs^i_t).
\end{equation}

To improve the physical plausibility of the scenarios, we discourage collisions between adversarial agents through a thresholded repulsive potential $\phi_{col}^{adv}$. The threshold $\tau$ ensures that the potential is active only when a pair of agents are closer than a safety margin. Furthermore, we find it sufficient to apply the repulsive potential $\phi_{col}^{adv}$ to the closest pair of adversarial agents. It is hence defined as:
\begin{equation}
    \phi_{col}^{adv}(\cS) = 
        - \min\Big( \min_{i,j \in \left\{1, ..., N\right\}, \,\, t \in \left\{0, ..., T\right\}}  d(\bs^i_t,\bs^j_t), \tau \Big).
\end{equation}

Finally, we regularize adversarial agents to prevent deviations from drivable areas with a repulsive potential $\phi_{dev}^{adv}$. This is applied between the adversarial agents and the off-road areas as defined by the map $\mathcal{M}$. We use a Gaussian potential $g(\bs^i_t, \cM)$ corresponding to the $i^{\text{th}}$ adversarial agent, which is applied across all timesteps and agents:
\begin{equation}
    \phi_{dev}^{adv}(\cS) = \sum_{i=0}^{N} \sum_{t=0}^T g(\bs^i_t, \cM).
\end{equation}
Additional details regarding the cost functions (such as the specific parameter values) are provided in the supplementary. Note that the realism of the generated scenarios is determined by the choice of regularizing terms in $\mathcal{C}$. While additional regularization may be beneficial, we find that the three terms in Eq. \eqref{eq:method:objective} are sufficient to find meaningful scenarios. We remark that our goal is to discover challenging scenarios that lie in the long tail of the distribution of traffic. Therefore, the scenarios discovered by our objective are not all likely to occur frequently in daily traffic. Importantly, however, the discovered scenarios are diverse, solvable, and enable learning more robust driving behaviors as demonstrated in \secref{sec:results:bbo} and \secref{sec:results:robustness}.

\begin{figure*}[t!]
	\centering
	\includegraphics[width=.99\textwidth]{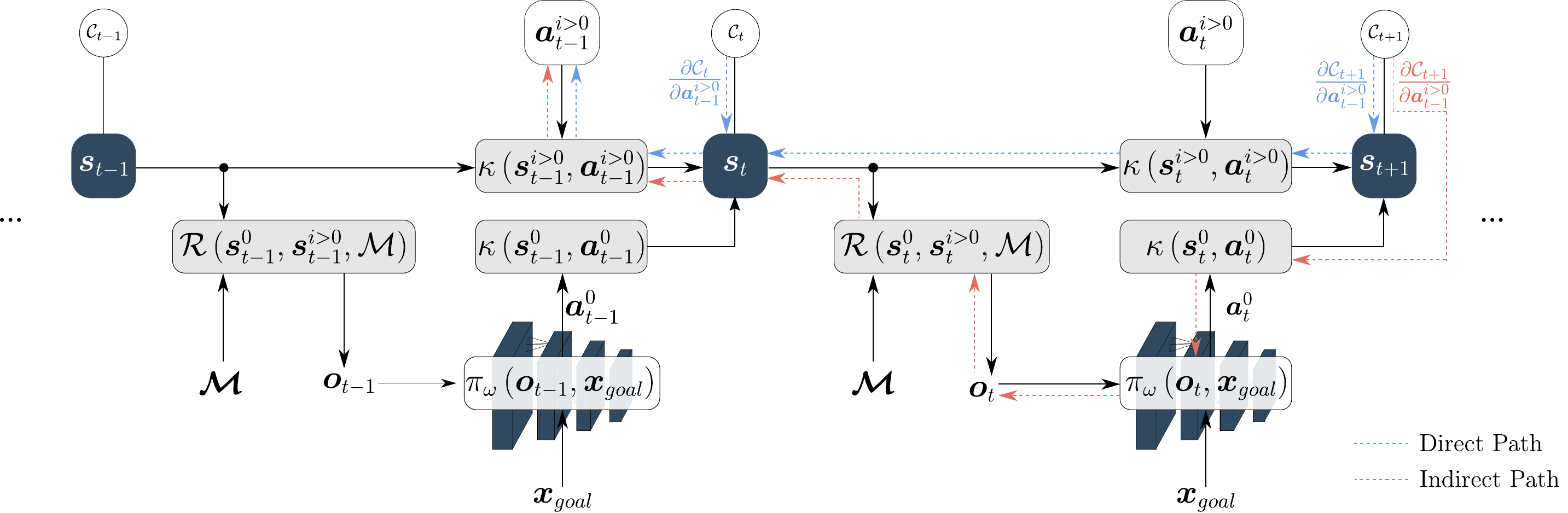}
    \caption{\textbf{Gradient paths.} To simulate a scenario, we render an observation $\bo_t$ for the driving policy $\pi_\omega$ under attack using a rendering function $\cR$. Both the driving policy and adversarial agents then take actions. The actions of the ego agent $\ba_t^{0}$ depend on the observation and a goal location $\bx_{goal}$. The actions of the adversarial agents $\ba_t^{i>0}$ are the parameters to optimize over to a safety-critical perturbation. Given the actions for all agents and current traffic state $\bs_t$, the next state $\bs_{t+1}$ is computed using a differentiable kinematics model $\kappa$. Gradients from the cost at time $t$ can then be propagated back to states at preceding timesteps. As shown, the derivative has components along two paths: an efficient \ellisblue{direct path} and a compute-intensive \ellisred{indirect path}.}
    \label{fig:method:gradient_paths}
\end{figure*}

\boldparagraph{Kinematics Gradients}
Given that the sequence of states $\cS$ is unrolled based on the differentiable kinematics model, we can backpropagate costs at any timestep $t$ to the set of actions $\left\{\ba_{t-1}, \ba_{t-2}, ..., \ba_{0}\right\}$ at previous timesteps. In the full unrolled computation graph of the simulation, the true gradients of the cost at any timestep can be taken \wrt the actions in preceding timesteps by recursively applying the chain rule along two paths: a direct path through the kinematics model and an indirect path, which additionally involves the driving policy $\pi_\omega$ and renderer $\mathcal{R}$. This is illustrated in \figref{fig:method:gradient_paths}.

With \methodName, we propose an approximation to the true gradients, which only considers the direct path and stops gradients through the indirect path. While this introduces an error in the gradient estimation, we empirically find it to work well while leading to several advantages. Firstly, as we show in \secref{sec:results:bbo}, it enables gradient-based generation in the common case where the rendering function or driving policy is non-differentiable, preventing gradients to be taken \wrt the indirect path. Secondly, even when all components are differentiable, taking gradients \wrt to the indirect path involves backpropagating through the driving policy and rendering function ({dotted red arrows} in \figref{fig:method:gradient_paths}), incurring significant computational overhead. We investigate this setting for \bevAgent{} where both the driving policy and rendering function are differentiable in \secref{sec:results:bbo} and show that given a fixed computational budget, this computational overhead leads to worse results compared to \methodName. We hypothesize that utilizing gradients through both paths becomes more important as the driving policy becomes robust to attacks.

\boldparagraph{Robust Training for \ac{il}}
\label{sec:method:robust}
After stress-testing the \ac{il}-based driving agents, we are further interested in improving robustness by augmenting the original training data with the generated safety-critical scenarios. To this end, we pursue a simple yet effective strategy: (1) we generate a large set of safety-critical scenarios, (2) we filter these for scenarios in which a privileged rule-based expert algorithm finds a safe alternate trajectory, (3) we collect a dataset of observation-waypoint pairs $\cD_{crit}$ for the filtered scenarios using the expert, and (4) we fine-tune the policy $\pi_\omega$ with the standard $L_1$ loss $\cL$ on a mix of the safety-critical data $\cD_{crit}$ and the original dataset $\cD_{reg}$:
\begin{equation}
    \omega^* = \argmin_{\omega} \nE_{(\bo_t, \bx_{goal}, \bw) \sim (\cD_{crit} \ \cup \ \cD_{reg}) } \left[ \cL (\bw, \pi_\omega \left(\bo_t, \bx_{goal} \right)) \right].
\end{equation}

%% file: sec_results.tex
\section{Experiments}
\label{sec:results}

We now present the research questions we aim to answer in our experimental study.

\noindent \textbf{Can gradient-based attacks outperform black-box optimization (\ac{bbo}) for safety-critical scenario generation?}
We are interested in reducing the optimization time needed to take a set of non-critical scenario initializations and find interesting scenarios. Given the computational overhead of computing gradients and performing a backward pass, we analyze the gains that can be achieved for this task with gradient-based attacks over \ac{bbo} in \secref{sec:results:bbo}. In addition, as shown in \figref{fig:method:gradient_paths}, there are two paths for gradients through a simulator. We aim to understand the computational cost of backpropagating through each path and the corresponding gains in terms of collision rates.

\noindent \textbf{Are gradient-based attacks applicable to non-differentiable simulators?} Our main experiments are conducted using a differentiable simulator that renders the BEV grid inputs for \bevAgent{}. In \secref{sec:results:transfuser}, we aim to investigate the applicability of \methodName{} to non-differentiable rendering functions, such as CARLA's camera and LiDAR sensors.

\noindent \textbf{Can we improve robustness by augmenting the training distribution with critical scenarios?} We are interested in the analyzing robustness of the fine-tuned \ac{il} model that uses the data augmentation strategy described in \secref{sec:method:robust}. In \secref{sec:results:robustness}, we investigate this on both the regular benchmark (hand-crafted scenarios) and held-out safety-critical test scenarios generated by \methodName{}. 

\subsection{Benchmarking IL Agents on Hand-Crafted Scenarios}
\label{sec:results:regular_eval}

To gain an initial understanding of their robustness, we first benchmark the agents used in our study with hand-crafted scenarios from CARLA\footnote{\url{https://leaderboard.carla.org/scenarios}}. As an additional benchmark that aims to maximize the traffic interactions achievable with such scenarios, we select a set of short routes through intersections involving dense traffic. We describe these benchmarks below. The results provide a reference for performance of our \bevAgent{} agent and the existing TransFuser agent on these settings which are relevant for the following experiments. All our experiments are conducted using \carla{} version 0.9.10.1.

\boldparagraph{Experimental Setup} \bevAgent{} and TransFuser~\cite{Prakash2021CVPR} are trained via supervised learning to imitate a privileged expert on data containing regular \carla{} traffic. The expert is a rule-based algorithm similar to the CARLA traffic manager autopilot\footnote{\url{https://carla.readthedocs.io/en/latest/adv_traffic_manager}}. We evaluate these models on two benchmarks: (1) the NEAT validation routes from \cite{Chitta2021ICCV}, and (2) a set of 82 routes through intersections in CARLA's Town10 with dense traffic. The NEAT routes provide a holistic evaluation of the driving performance, but the evaluation is time-consuming. This set contains routes varying in length from 100m to 3km with regular \carla{} traffic and hand-crafted scenarios. Since several of the routes are long and contain low traffic densities, poor collision avoidance has limited impact on the final metrics. For a more focused evaluation on collisions with traffic, the Town10 intersection routes are shorter in length (80m-100m). In this setting, we ensure a high density of dynamic agents by spawning vehicles at every possible spawn point permitted by the CARLA simulator. Furthermore, each route is guaranteed to contain a hand-crafted scenario in which multiple vehicles enter the intersection from different directions at the same time. We selected Town10 for this benchmark as we found it to be the most challenging in preliminary experiments.

\boldparagraph{Metrics} On both of these benchmarks, we report the official metrics of the CARLA leaderboard, \textbf{Route Completion (RC)}, \textbf{Infraction Score (IS)} and \textbf{Driving Score (DS)}. RC is the percentage of the route completed by an agent before it gets blocked or deviates from the route. IS is a cumulative multiplicative penalty for every red light violation, stop sign violation, collision, and lane infraction. DS is the final metric, computed as the RC multiplied by the IS for each route. Each model is tested with three different evaluation seeds. In addition, we report the \textbf{collision rate (CR)}, which is the percentage of routes in which the agent collided while traversing an intersection. Additional details regarding the driving metrics, rule-based expert, and training dataset for the driving policy are provided in the supplementary material.

\boldparagraph{Results} The performance of the two driving agents as well as the rule-based expert which uses privileged information is shown in \tabref{tab:neat}. Note that the three methods have different inputs, and are not directly comparable. \bevAgent{} achieves a superior IS and DS in comparison to TransFuser. In particular, its significantly higher IS on the NEAT routes indicates that it is proficient at avoiding collisions when placed in sparse and non-adversarial \carla{} traffic. On the Town10 intersections, \bevAgent{} has a better IS than TransFuser, but we observe that the CR of both agents is similar (17.48\%). This is much higher than the expert (CR=3.66\%), showing that hand-crafted scenarios in dense traffic remain challenging for current \ac{il}-based methods. These hand-crafted scenarios are not adaptive to the agent under test, \ie, the same scenarios are applied for both \bevAgent{} and TransFuser. In the following, we study the more targeted approach of actively generating safety-critical scenarios that are adaptive to the agent being attacked.

\begin{table}[t!]
    \centering
    \setlength{\tabcolsep}{0.015\textwidth}
    \resizebox{\textwidth}{!}{
        \begin{tabular}{l|rrrr|rrrr}
            \toprule
            & \multicolumn{4}{c|}{\textbf{NEAT validation routes}~\cite{Chitta2021ICCV}} & \multicolumn{4}{c}{\textbf{Town10 intersections}} \\
            \midrule
            \textbf{Method} & \multicolumn{1}{c}{\textbf{RC} $\uparrow$}& \multicolumn{1}{c}{\textbf{IS} $\uparrow$}&\multicolumn{1}{c}{ \textbf{DS} $\uparrow$ }  & \multicolumn{1}{c|}{\textbf{CR} $\downarrow$} & \multicolumn{1}{c}{\textbf{RC} $\uparrow$}& \multicolumn{1}{c}{\textbf{IS} $\uparrow$}&\multicolumn{1}{c}{ \textbf{DS} $\uparrow$} & \multicolumn{1}{c}{\textbf{CR} $\downarrow$}\\
            \midrule
            \bevAgent & 96.77\scriptsize{$\pm$3.32} & {0.95\scriptsize{$\pm$0.00}} & {92.24\scriptsize{$\pm$3.32}}& {2.38\scriptsize{$\pm$4.12}} &
            {93.86\scriptsize{$\pm$0.14}} &
            {0.92\scriptsize{$\pm$0.01}} &
            {86.74\scriptsize{$\pm$0.67}} &  {17.48\scriptsize{$\pm$1.86}} \\
            TransFuser~\cite{Prakash2021CVPR}  & {99.25\scriptsize{$\pm$1.30}} & 0.78\scriptsize{$\pm$0.03} & 77.59\scriptsize{$\pm$2.01}& {11.90\scriptsize{$\pm$4.12}}&
            93.68\scriptsize{$\pm$2.01}&
            0.85\scriptsize{$\pm$0.00}&
            80.03\scriptsize{$\pm$0.79}&
            {17.48\scriptsize{$\pm$0.70}}\\
            \midrule
            Privileged Expert  & 99.83\scriptsize{$\pm$0.07} & 1.00\scriptsize{$\pm$0.00} &  99.83\scriptsize{$\pm$0.07}& {0.00\scriptsize{$\pm$0.00}}&
            94.89\scriptsize{$\pm$0.33}&
            0.97\scriptsize{$\pm$0.00}&
            92.81\scriptsize{$\pm$0.53}&
            3.66\scriptsize{$\pm$0.00} \\
            \bottomrule
        \end{tabular}
    }
    \vspace{0.15cm}
    \caption{\textbf{Performance on hand-crafted scenarios.} We show the mean $\pm$ std over 3 evaluations. \bevAgent{} has fewer infractions than TransFuser on the NEAT validation routes. However, both agents collide in over 17\% of the Town10 intersection routes.}
    \label{tab:neat}
\end{table}

\subsection{Comparison to BBO for Safety-Critical Scenario Generation}
\label{sec:results:bbo}
Next, we analyze the efficacy of \methodName{} for the generation of safety-critical scenarios, by comparing it with several \ac{bbo} baselines for attacking \bevAgent{}.

\boldparagraph{Experimental Setup} One scenario in our experimental setup involves rolling out a policy for 20 seconds of simulation time (80 timesteps at 4fps). We find this time horizon to be sufficient for the ego agent to traverse a route from the start location to the end location while coming in close proximity to the adversarial agents. We compare several adversarial optimization techniques on 80 such scenarios. We obtain 4 maps (Town03-Town06) from the \carla{} simulator. The 4 maps have a wide variety of road layouts, including intersections, single-lane roads, multi-lane highways, exits, and roundabouts (additional details in supplementary). We sample a dense set of candidate start locations and end locations for the ego agent from the set of all junctions available in these 4 maps. The 80 ego agent routes in our evaluation are obtained by uniformly sampling 20 candidate routes per \carla{} town.

\boldparagraph{Initializing Background Traffic} For each ego agent route in our evaluation, we now aim to retrieve initial routes for the adversarial agents that are in the direct surroundings of the ego agent. To this end, we retrieve all potential routes from the dense set of candidate start and end locations that closely pass by the ego agent's route. These are assigned as the corresponding start and end locations for the adversarial agents in that scenario. We use the privileged expert to drive the adversarial agents along their assigned route. This yields a non-critical initial scenario that mimics the CARLA traffic and allows explicit control over the number of adversarial agents involved. We use three traffic densities in our evaluation: 1 agent, 2 agents, and 4 agents. For more details, please refer to the supplementary material.

\boldparagraph{Metrics} We evaluate the adversarial scenarios using the \textbf{collision rate (CR)}, which is the percentage of routes for which the adversarial scenario search yielded a collision while respecting behavioral constraints. In particular, a search is only considered successful if all adversarial agents stay on drivable parts of the map (\ie, the road) and do not collide with other adversarial agents. To evaluate the convergence, we report the average \textbf{time to 50\% collision rate ($t_{50\%}$)}. This measures the average computation cost (in GPU seconds) required to find a collision in 50\% of the total scenarios available. Finally, we report the runtime of each technique as the average number of optimization \textbf{seconds per iteration (s/it)}. The $t_{50\%}$ and s/it metrics for \methodName{} as well as all baselines are evaluated on a single RTX 2080Ti GPU. For all methods, we use a compute budget of 180 seconds per route on a single GPU, leading to a total experimental budget of up to 4 GPU hours for 80 routes.

\boldparagraph{Results} We now assess the efficiency of \methodName{} compared to \ac{bbo}. To this end, we report the CR, $t_{50\%}$ and s/it of our approach and several baselines in \tabref{tab:num_agents}. We consider the three traffic density settings separately, as well as the overall metrics for the complete set of 80$\times$3 scenarios. Our baselines optimize the scenario parameters via \ac{bbo}. In particular, besides \textbf{Random Search} and \textbf{Bayesian Optimization}, we consider \textbf{SimBA}~\cite{Guo2019ICML}, \textbf{CMA-ES}~\cite{Hansen2001ecj} and \textbf{Bandit-TD}~\cite{Ilyas2019ICLR}. SimBA is a variant of Random Search that greedily maximizes the objective and CMA-ES is a state-of-the-art evolutionary algorithm. Finally, Bandit-TD computes numerical gradients by integrating priors into a finite differences approach.

\begin{table}[t!]
    \centering
    \setlength{\tabcolsep}{0.015\textwidth}
    \resizebox{\linewidth}{!}{
        \begin{tabular}{l|rrr|rrr|rrr|rrr}
             \toprule
             & \multicolumn{3}{c|}{\textbf{1 Agent}} & \multicolumn{3}{c|}{\textbf{2 Agents}} &\multicolumn{3}{c|}{\textbf{4 Agents}} &\multicolumn{3}{c}{\textbf{Overall}} \\
             \midrule
             \textbf{Method} & \textbf{CR} $\uparrow$& \textbf{$t_{50\%}$} $\downarrow$ & \textbf{s/it} $\downarrow$
             & \textbf{CR} $\uparrow$& \textbf{$t_{50\%}$} $\downarrow$ & \textbf{s/it} $\downarrow$
             & \textbf{CR} $\uparrow$& \textbf{$t_{50\%}$} $\downarrow$ & \textbf{s/it} $\downarrow$
             & \textbf{CR} $\uparrow$& \textbf{$t_{50\%}$} $\downarrow$ & \textbf{s/it} $\downarrow$\\
             \midrule
             Random Search & 62.50 & \textbf{9.25} & 1.30 & 68.75 & 7.38 & 1.35 & 68.75 & 15.22  & 1.48 & 66.67 & 9.66 & 1.38\\
             Bayesian Optimization  & 63.75 & 11.88 & 1.46 & 68.75 & 10.01 & 1.66 & 63.75 & 22.12 & 2.06 & 65.00 & 14.34 & 1.73\\  
             SimBA~\cite{Guo2019ICML}  & 60.00 & 14.14 & 1.30 & 71.25 & 14.35 & 1.35 & 61.25 & 19.68  & 1.48 & 64.17 & 15.84 & 1.38\\
             CMA-ES~\cite{Hansen2001ecj}  & 67.50 & 9.34 & 1.31 & 75.00 & \textbf{6.73} & 1.36 & 62.50 & 9.39  & 1.52 & 68.33 & 8.17 & 1.40\\
             Bandit-TD~\cite{Ilyas2019ICLR} & 37.50 & \multicolumn{1}{c}{\hspace{3mm}-} & 3.87 & 30.00 & \multicolumn{1}{c}{\hspace{3mm}-}  & 4.39 & 21.25 & \multicolumn{1}{c}{\hspace{3mm}-}  &5.02 & 29.58 & \multicolumn{1}{c}{\hspace{3mm}-}  & 4.43 \\
              \midrule
             \methodName{} Direct + Indirect & 78.75 & 19.33 & 3.17 & 72.50 & 14.68 & 3.25 & 76.25 & 14.67 & 3.40  & 75.83 & 16.14 & 3.27\\
             \methodName{} (Ours) & \textbf{86.25} & 9.98 & 1.78 & \textbf{82.50} & 6.96 & 1.88 & \textbf{78.75} & \textbf{6.40} & 2.03 & \textbf{82.50} &\textbf{7.78} & 1.90\\
             \bottomrule
        \end{tabular}
    }
    \vspace{0.15cm}
    \caption{\textbf{Critical scenario generation on CARLA.} We show the mean CR, $t_{50\%}$ and s/it for different optimization techniques in three traffic settings, as well as the aggregated metrics. \methodName{} finds collisions in over 80\% of the initializations, significantly outperforming all baselines. Using only the direct path (Ours) leads to the highest CR and is faster than using gradients from both the direct and indirect paths.}
    \label{tab:num_agents}
\end{table}

\methodName{} obtains a significantly higher CR than the \ac{bbo} baselines in all 3 settings, increasing the number of scenarios for which a safety-critical perturbation is found by over 20\%. Among the \ac{bbo} baselines, CMA-ES attains the best overall scores with respect to both CR and $t_{50\%}$. Interestingly, the best performance for \ac{bbo} is often observed for $N=2$ agents. As we increase $N$ from 1 to 2, it becomes easier for the baselines to find one nearby agent that can be perturbed to collide with the ego agent. However, further increasing $N$ to 4 makes it harder to maintain plausible trajectories where the adversarial agents do not collide with each other or go off-road, leading to reduced performance. As the dimensionality of the search space increases (\eg $N=4$), \methodName{} begins to outperform the baselines in terms of $t_{50\%}$ by a large margin.

We also compare the proposed approximation in \methodName{} against the setting where we use gradients through entire simulation, including the driving policy and renderer (``\methodName{} Direct + Indirect" in \tabref{tab:num_agents}). While also reliably finding safety-critical perturbations, the computational overhead of backpropagating through the indirect path leads to worse results given the same computation budget. This suggests the approximation in \methodName{} is reasonable for efficiently generating safety-critical scenarios. Additional results and details regarding the hyper-parameter choices for \ac{bbo} are provided in the supplementary material. Since we observe that gradients through the direct path only are sufficient, we now conduct a detailed qualitative analysis where we apply \methodName{} to attack TransFuser, which requires the use of CARLA's non-differentiable camera and LiDAR sensors for rendering.

\begin{figure}[t!]
    \centering
    \hspace{-0.5cm}
    \includegraphics[width=\columnwidth]{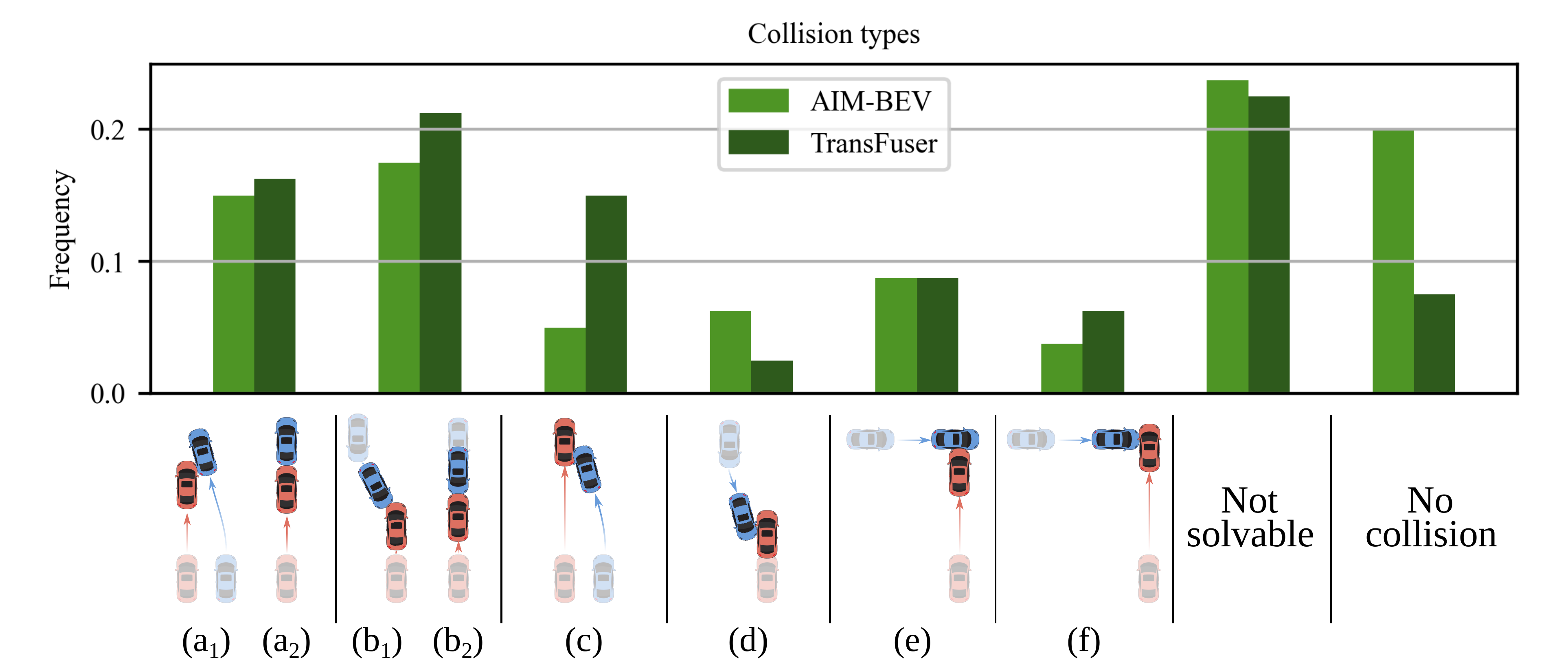}
    \caption{\textbf{Collision types.} For a traffic density of 4 agents, we observe that \methodName{} generates a diverse set of challenging but solvable scenarios. We group these into 6 clusters (a-f). The cluster illustrations depict the ego agent in \ellisred{red} and the adversarial agent in \ellisblue{blue}. The scenarios include (a) cut-ins ahead of the ego agent and rear-ends caused by the ego agent, (b) head-on collisions, (c) merges, (d) side collisions with oncoming traffic, and t-bone collisions in intersections (e and f).}
    \label{fig:results:collision_types}
\end{figure}

\begin{figure}[t!]
	\centering
	\begin{subfigure}{0.49\linewidth}
		\centering
        \setlength{\tabcolsep}{0.0\textwidth}
		\begin{tabular}{c c}
		     \includegraphics[width=0.495\linewidth]{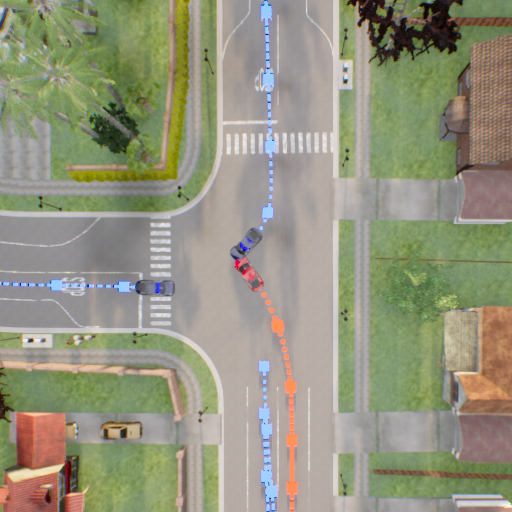} &
		     \includegraphics[width=0.495\linewidth]{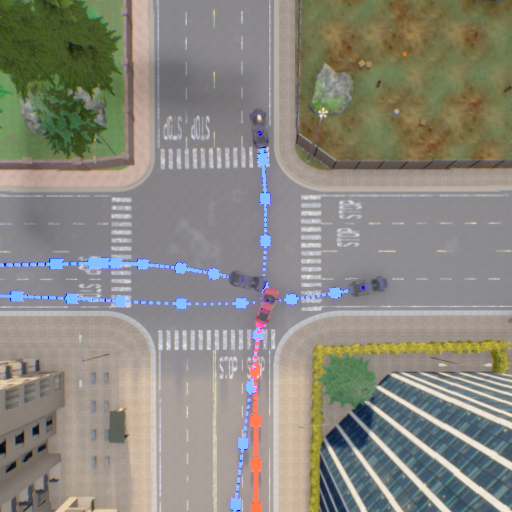}
		\end{tabular}
		\label{subfig:results:attacks_qualitative_examples:aimbev}
		\caption{We show two scenarios generated by \methodName{} in which \bevAgent{} enters an intersection and fails to yield to the perturbed background traffic. This leads to t-bone collisions, either by the ego agent (left) or the adversarial agent (right), corresponding to clusters (e) and (f) in \figref{fig:results:collision_types}.}
	\end{subfigure}\hfill
	\begin{subfigure}{0.49\linewidth}
		\centering
        \setlength{\tabcolsep}{0.0\textwidth}
		\begin{tabular}{c c}
		     \includegraphics[width=0.995\linewidth]{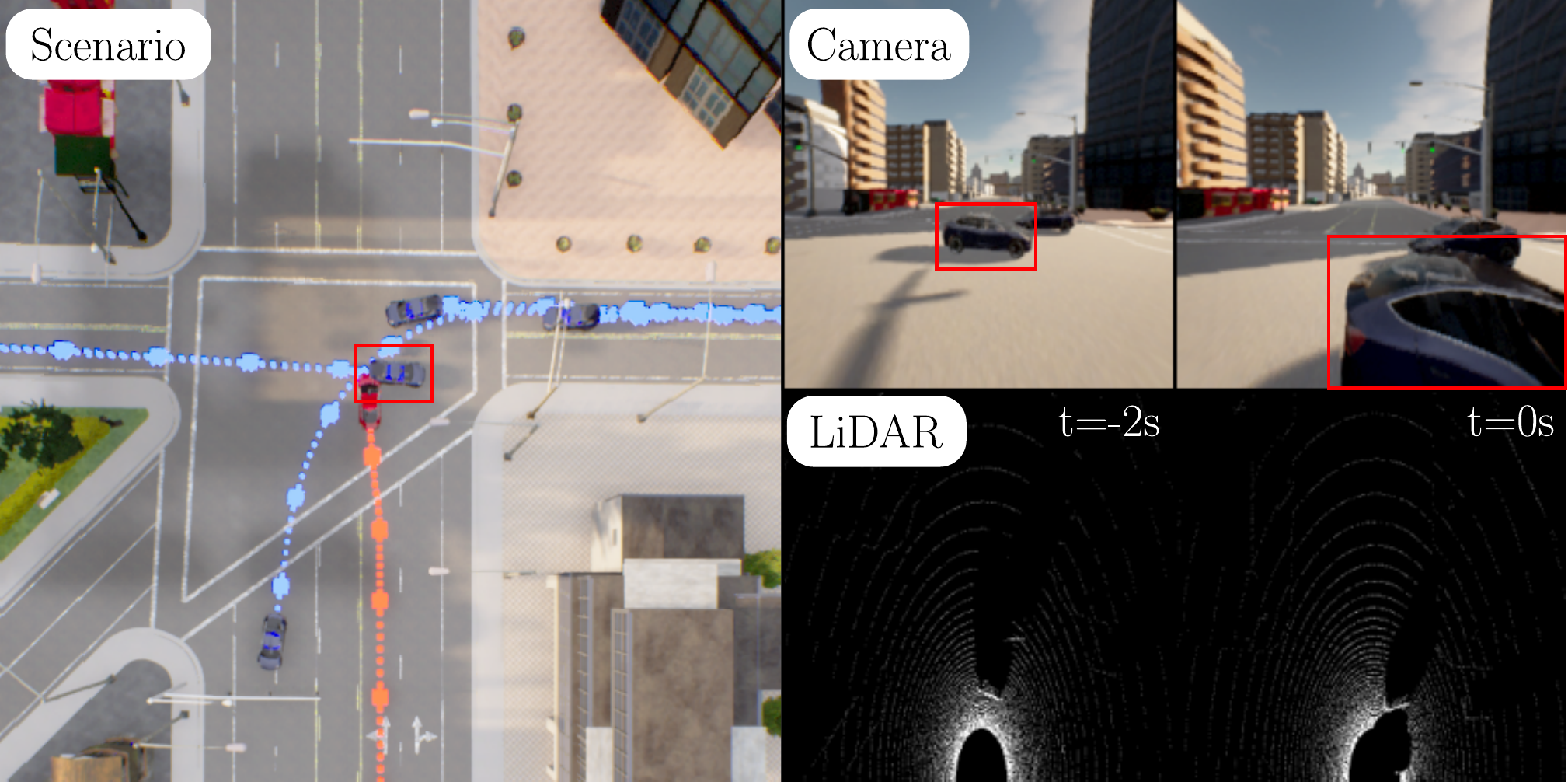}
		\end{tabular}
		\label{subfig:results:attacks_qualitative_examples:transfuser}
		\caption{We show a scenario along with camera and LiDAR inputs two seconds and zero seconds before the safety-critical situation for TransFuser~\cite{Prakash2021CVPR}. The model is unable to slow down to prevent a collision with the adversarial agent which stops inside the intersection ({\color{red}red box}).}
	\end{subfigure}
	\caption{\textbf{Qualitative examples of safety-critical scenarios generated by \methodName{}.} Ego agent in \ellisred{red}, adversarial agent in \ellisblue{blue}. Best viewed zoomed in.}
	\label{fig:results:attacks_qualitative_examples}
\end{figure}

\subsection{Analysis of Safety-Critical Scenarios}
\label{sec:results:transfuser}
In this section, we analyze the safety-critical scenarios generated by \methodName{} for both \bevAgent{} and TransFuser in detail. Specifically, we show the distribution of the resulting scenarios with a traffic density of $N=4$ agents in \figref{fig:results:collision_types}. 
For both driving agents, we first filter out the set of scenarios where \methodName{} is unable to find a collision (``No Collision") as well as those that are not solvable by the rule-based expert (``Not Solvable"). We cluster the remaining scenarios using k-means (similar to \cite{Rempe2021ARXIV}) to obtain 6 clusters of failure modes such as cut-ins ($\text{a}_1$), rear-ends ($\text{a}_2$) and unsafe behavior in unprotected turns (e,f). From the frequency of scenarios with ``No collision" in \figref{fig:results:collision_types}, we observe that both \bevAgent{} and TransFuser collide in at least 80\% of the scenarios. This is a significant deviation from the collision avoidance of both models in the benchmarks shown in \tabref{tab:neat}, where they attain a CR below 20\%. The large amount of collisions for TransFuser indicates that \methodName{} can achieve promising results when applied out-of-the-box to driving simulators with non-differentiable rendering functions.

We show qualitative examples in \figref{fig:results:attacks_qualitative_examples}, and additional examples in the supplementary material. Both \bevAgent{} and TransFuser frequently collide in intersections when they encounter traffic that behaves differently from the traffic observed during training. Importantly, the ``Not solvable" column shows that for both agents, only around 20\% of the scenarios have no feasible alternate trajectory. This leaves a large proportion of solvable scenarios in the 6 clusters shown in \figref{fig:results:collision_types}. The most frequent failure modes of both models are observed in clusters (a) and (b), which involve cut-ins, rear-ends, and head-on collisions. The rule-based expert solves these challenging scenarios by accurately forecasting the motion of the adversarial actors using privileged information. Interestingly, the failure cases are fairly evenly distributed over the 6 clusters which involve a wide variety of relative orientations between the colliding agents. The examples in \figref{fig:results:collision_types} correspond to clusters (e) and (f). We highlight examples from clusters (a$_1$) and (a$_2$) for our experiment in \figref{fig:results:aimbev_king_qual_examples}. The high frequency and diversity of solvable scenarios generated by \methodName{} indicate its potential to augment the original training data for \ac{il} models, which we investigate next.

\subsection{Evaluating Robustness after Fine-Tuning}
\label{sec:results:robustness}
Finally, we analyze the efficacy of the generated scenarios in augmenting the original training distribution to yield more robust driving agents. Here, we evaluate robustness both with respect to safety-critical scenarios generated by \methodName{} and to hand-crafted scenarios in the CARLA simulator (using the Town10 intersections benchmark).

\boldparagraph{Experimental Setup} The goal of this experiment is to collect training data for improving collision avoidance. To this end, we build a large set of safety-critical scenarios by attacking \bevAgent{} using initializations from Town03-Town06 of CARLA with $N=4$ agents. To ensure meaningful supervision, we filter the resulting scenarios for ones where \methodName{} finds collisions that are solvable by the expert. This results in around 300 scenarios from which we hold out 20\% for evaluation. We ensure that there is no overlap between the training and evaluation during this split by preventing routes with the same ego vehicle start location from being in both splits. Additional details regarding the training data and hyper-parameters are provided in the supplementary material.

\begin{table}[t!]
    \centering
    \setlength{\tabcolsep}{0.025\textwidth}
    \resizebox{\textwidth}{!}{
        \begin{tabular}{l|c|cccc}
            \toprule
            & \multicolumn{1}{c|}{\textbf{Held-out \methodName{} scenarios}} & \multicolumn{4}{c}{\textbf{Hand-crafted scenarios (Town10 intersections)}} \\
            \midrule
            \textbf{Dataset} & \textbf{CR} $\downarrow$ & \textbf{RC} $\uparrow$& \textbf{IS} $\uparrow$& \textbf{DS} $\uparrow$ & \textbf{CR} $\downarrow$\\
            \midrule
            No Fine-tuning & {100.00\scriptsize{$\pm$0.00}} & {93.86\scriptsize{$\pm$0.14}} &
            {0.92\scriptsize{$\pm$0.01}} &
            {86.74\scriptsize{$\pm$0.67}} &  {17.48\scriptsize{$\pm$1.86}} \\
            $\cD_{reg}$ & {57.14\scriptsize{$\pm$0.00}} & \textbf{95.66\scriptsize{$\pm$0.51}} & 0.90\scriptsize{$\pm$0.00} & 86.85\scriptsize{$\pm$0.62} & 19.51\scriptsize{$\pm$0.00}  \\
            $\cD_{crit}$ & \textbf{28.57\scriptsize{$\pm$0.00}} & 91.92\scriptsize{$\pm$0.19}&\textbf{0.96\scriptsize{$\pm$0.00}}&88.37\scriptsize{$\pm$0.41}&\textbf{6.10\scriptsize{$\pm$0.00}}\\
            $\cD_{crit} \ \cup \ \cD_{reg}$ & \textbf{28.57\scriptsize{$\pm$0.00}} & {94.42\scriptsize{$\pm$0.36}} & \textbf{0.96\scriptsize{$\pm$0.36}} & \textbf{90.20\scriptsize{$\pm$0.00}} & {8.13\scriptsize{$\pm$0.70}} \\
            \bottomrule
        \end{tabular}
    }
    \vspace{0.15cm}
    \caption{\textbf{Robust training for \bevAgent{}.} Results shown are the mean and std over 3 evaluation seeds. Fine-tuning with safety-critical scenarios reduces the CR by over 50\%  on other safety-critical scenarios as well as hand-crafted scenarios from \carla{}.}
    \label{tab:results:robust_il}
\end{table}

\begin{figure}[t!]
	\centering
	\begin{subfigure}{0.49\linewidth}
		\centering
        \setlength{\tabcolsep}{0.0\textwidth}
		\begin{tabular}{c c}
		     \includegraphics[width=0.495\linewidth]{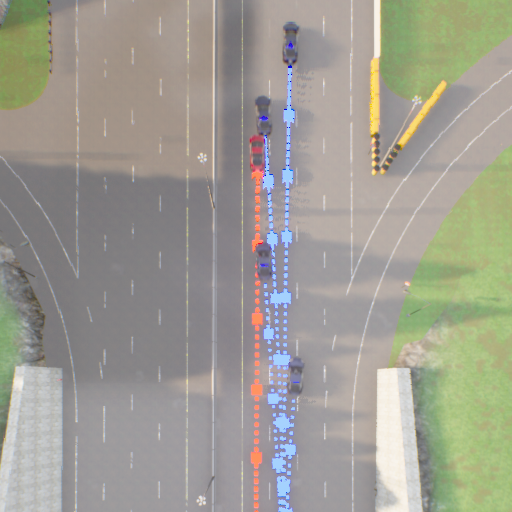} &
		     \includegraphics[width=0.495\linewidth]{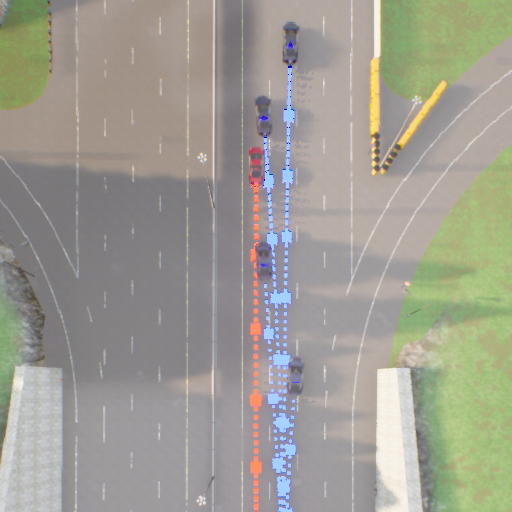}
		\end{tabular}
		\label{subfig:results:aimbev_king_qual_examples:king}
		\caption{Maintaining a safe distance during a merge.}
	\end{subfigure}\hfill
	\begin{subfigure}{0.49\linewidth}
		\centering
        \setlength{\tabcolsep}{0.0\textwidth}
		\begin{tabular}{c c}
		     \includegraphics[width=0.495\linewidth]{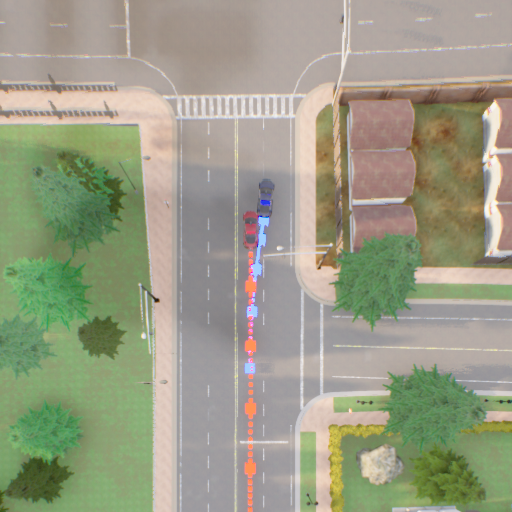} &
		     \includegraphics[width=0.495\linewidth]{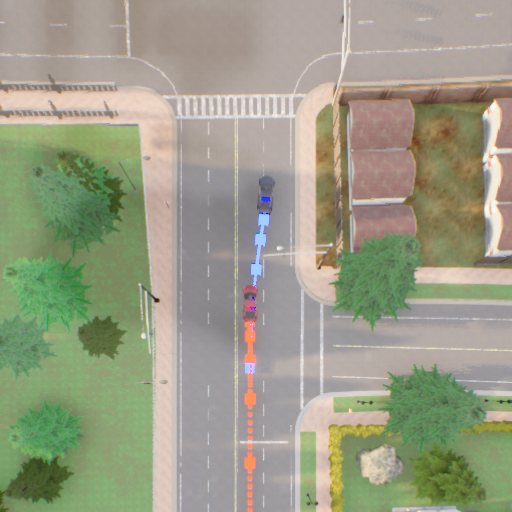}
		\end{tabular}
		\label{subfig:results:aimbev_king_qual_examples:zerosplit}
		\caption{Slowing down to avoid a rear-end.}
	\end{subfigure}
	\caption{\textbf{Improved collision avoidance on held-out \methodName{} scenarios with \bevAgent{}.} Comparison of the original model (left) vs. the robust model fine-tuned on $\cD_{crit} \ \cup \ \cD_{reg}$ (right). Ego agent in \ellisred{red}, adversarial agent in \ellisblue{blue}. Best viewed zoomed in.}
	\label{fig:results:aimbev_king_qual_examples}
\end{figure}

\boldparagraph{Results} We report the driving performance of \bevAgent{} after fine-tuning on $\cD_{crit} \ \cup \ \cD_{reg}$ in \tabref{tab:results:robust_il}. Since the trajectories of the adversarial agents are fixed after optimization via \methodName{}, some of the scenarios may be solvable by simply adopting different overall driving styles, rather than becoming more proficient at collision avoidance. To quantify this, we fine-tune each model with only the original training data $\cD_{reg}$ as a baseline, which reduces the CR from 100\% to 57.14\% on the held-out KING scenarios. Additionally, we compare to fine-tuning on only the critical scenarios $\cD_{crit}$ and the initial checkpoint from \tabref{tab:neat} (``No Fine-tuning"). Among the three fine-tuning strategies, using only $\cD_{reg}$ leads to unsatisfactory results, with a CR of 19.51\% on the Town10 intersections benchmark. Using only $\cD_{crit}$ leads to a large reduction in CR on both evaluation settings. However, the model has a lower RC and only a small improvement in DS when compared to the $\cD_{reg}$ baseline on the Town10 intersections. Finally, using the combined dataset of $\cD_{crit} \ \cup \ \cD_{reg}$ gives the best results. In this setting, we obtain a CR of 28.57\% on the KING scenarios, which is identical to the model fine-tuned with only $\cD_{crit}$. However, the DS of this model on Town10 is improved by over 3 points, since it reduces the CR while maintaining a similar RC to the original model. This shows that the simple strategy of fine-tuning on a mixture of regular and safety-critical data is an effective way of learning from the scenarios generated by \methodName{}.

In \figref{fig:results:aimbev_king_qual_examples}, we show qualitative driving examples of the original and fine-tuned \bevAgent{} agents on held-out \methodName{} scenarios, which belong to clusters (a$_1$) and (a$_2$) from \figref{fig:results:collision_types}. While these scenarios are straightforward to handle for an expert driver, \bevAgent{} fails to brake for a vehicle stopping in between two lanes and is unable to maintain a safe distance in merging maneuvers, which highlights its brittleness in o.o.d scenarios. These scenario types do not frequently emerge naturally from the CARLA simulator's background agent behavior which governs $\cD_{reg}$. By incorporating data from $\cD_{crit}$ during training, the driving agent can learn to handle these scenarios safely.

%% file: sec_conclusion.tex
\section{Conclusion}
We make substantial advances toward the generation of safety-critical traffic scenarios. We propose a novel gradient-based generation procedure, \methodName{}, which achieves significantly higher success rates compared to existing \ac{bbo}-based approaches while being more efficient. The key to our success is a compute-efficient direct gradient path through a kinematic motion model to guide the adversarial scenario generation process. Our analysis indicates that our method can achieve promising results when applied out-of-the-box to arbitrary driving agents. Furthermore, we show that despite having access to privileged \ac{bev} semantic maps as inputs, state-of-the-art \ac{il}-based driving policies are surprisingly brittle to minor perturbations in the behavior of the background actors. By augmenting their training data with scenarios from \methodName{}, we are able to significantly improve their collision avoidance. Exploring the robustness of agents with different training procedures (\eg \ac{rl}) offers an interesting direction for future research.